\documentclass{article}

\usepackage{hyperref}
\usepackage{multirow}
\usepackage{pifont}
\usepackage{graphicx}
\usepackage{wrapfig}
\usepackage{caption}
\usepackage{subcaption}
\usepackage[preprint]{neurips_2024}

\usepackage[utf8]{inputenc} % allow utf-8 input
\usepackage[T1]{fontenc}    % use 8-bit T1 fonts
\usepackage{hyperref}       % hyperlinks
\usepackage{url}            % simple URL typesetting
\usepackage{booktabs}       % professional-quality tables
\usepackage{amsfonts}       % blackboard math symbols
\usepackage{nicefrac}       % compact symbols for 1/2, etc.
\usepackage{microtype}      % microtypography
\usepackage{xcolor}         % colors

\title{RepGhost: A Hardware-Efficient Ghost Module via Re-parameterization}

\author{%
  Chengpeng Chen\thanks{Joint first authors.} , \ Zichao Guo\footnotemark[1] \thanks{Corresponding author.} , \ Haien Zeng, \ Pengfei Xiong, \ Jian Dong \\ 
  Shopee
}

\begin{document}

\maketitle

\begin{abstract}
   Feature reuse has been a key technique in light-weight convolutional neural networks (CNNs) architecture design. Current methods usually utilize a concatenation operator to keep large channel numbers cheaply (thus large network capacity) by reusing feature maps from other layers. Although concatenation is parameters- and FLOPs-free, its computational cost on hardware devices is non-negligible. To address this, this paper provides a new perspective to realize feature reuse implicitly and more efficiently instead of concatenation. A novel hardware-efficient RepGhost module is proposed for implicit feature reuse via re-parameterization, instead of using concatenation operator. Based on the RepGhost module, we develop our efficient RepGhost bottleneck and RepGhostNet. Experiments on ImageNet and COCO benchmarks demonstrate that our RepGhostNet is much more effective and efficient than GhostNet and MobileNetV3 on mobile devices. Specially, our RepGhostNet surpasses GhostNet 0.5$\times$ by 2.5\% Top-1 accuracy on ImageNet dataset with less parameters and comparable latency on an ARM-based mobile device. Code and model weights are available at \url{https://github.com/ChengpengChen/RepGhost}.
\end{abstract}

\section{Introduction}
\label{sec:intro}

\begin{figure}[t]
  \centering
%   \fbox{\rule{0pt}{2in} \rule{0.9\linewidth}{0pt}}
   \includegraphics[trim=0cm 0cm 0cm 0cm, clip, width=0.65\linewidth]{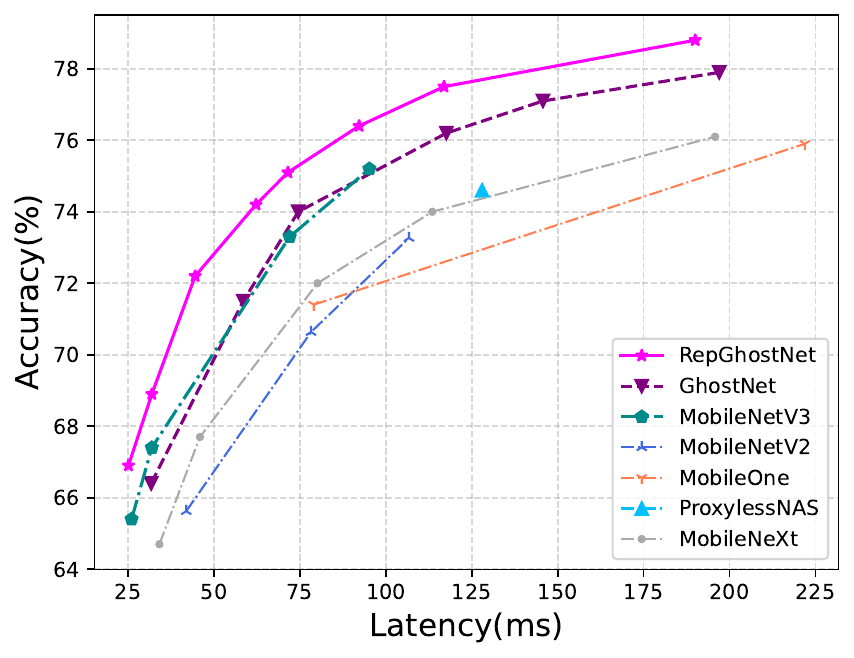}
   \vspace{-.25cm}
   \caption{Top-1 accuracy on ImageNet dataset \textbf{vs.} latency on an ARM-based mobile device, refer to Section~\ref{sec:exp} for the detail and Appendix~\ref{sec:more-latency-eval} for more running devices.}
   \label{fig:acc-to-latency}
   % \vspace{-.5cm}
\end{figure}

With the mobile and portable devices being ubiquitous, efficient convolutional neural networks (CNNs) are becoming indispensable due to the limited computational resources. Different methods are proposed to achieve the same goal of making CNNs more efficient in recent years, such as light-weight architecture design~\cite{ma2018shufflenet,howard2017mobilenets,sandler2018mobilenetv2,zhou2020rethinking,chen2021dynamic}, neural architecture search~\cite{tan2019mnasnet,kim2022proxyless,guo2020single,wu2019fbnet}, pruning~\cite{liu2019metapruning,molchanov2016pruning,anwar2017structured,he2017channel}, etc, and a lot of progress has been made.

In the field of CNNs architecture design, large channel numbers often means large network capacity~\cite{he2016deep,huang2017densely}, especially for light-weight CNNs~\cite{ma2018shufflenet,howard2019searching}. As stated in~\cite{ma2018shufflenet}, given a fixed floating-point operations (FLOPs), the light-weight CNNs prefer to use \textit{spare connections} ($e.g.$, group or depthwise convolution) and \textit{feature reuse}. They both have been well-explored and many representative light-weight CNNs have been proposed in recent years~\cite{howard2017mobilenets,howard2019searching,zhang2018shufflenet,ma2018shufflenet,zhou2020rethinking,cholletxception}. Spare connections are designed to keep large network capacity with low FLOPs, while feature reuse aims to \textit{explicitly} keep large number of channel by simply preserving existing features from different layers, which is often achieved by concatenation operations along channel dimension~\cite{huang2017densely,han2020ghostnet,szegedy2015going}. For example, in DenseNet, feature maps from previous layers are reused and sent to their subsequent layers within a stage, resulting in more and more channels. GhostNet proposes to generate more feature maps from cheap operations, and concatenate them with original ones for keeping large number of channels.
ShuffleNetV2 processes only half of channels and keeps the other half to be concatenated. They all utilize the feature reuse approach via concatenation to enlarge channel numbers while keeping FLOPs low. It seems that concatenation has been a standard and elegant operation for feature reuse, since it is parameters- and FLOPs-free. 

However, parameters and FLOPs are not direct cost indicators for actual runtime performance of machine learning models~\cite{dehghani2021efficiency,ma2018shufflenet}. Although concatenation operation is parameters- and FLOPs-free, its computational cost on hardware devices is non-negligible. To verify this, we provide detailed analysis in Section~\ref{subsec:feature-reuse} and find that concatenation operation is much more inefficient than add operation on hardware devices due to its complicated memory copy process. Therefore, it is noteworthy to explore a better and more hardware-efficient way for feature reuse beyond concatenation operation.

Recently, structural re-parameterization has proved its effectiveness in CNNs architecture design, including ExpandNets~\cite{guo2020expandnets}, ACNet~\cite{ding2019acnet}, and RepVGG~\cite{ding2021repvgg}. It converts complex training-time architectures into simpler inference-time ones equivalently without any extra inference costs. Inspired by this, we propose to utilize structural re-parameterization, instead of the widely-used concatenation, to realize feature reuse \textit{implicitly} for hardware-efficient architecture design.

In this paper, we propose a hardware-efficient RepGhost module via structural re-parameterization to realize feature reuse implicitly.
Note that it is not just to apply re-parameterization technique to Ghost module, but to design our novel and efficient module for fast inference. To be specific, we first remove the inefficient concatenation operator, and then modify the architecture to satisfy the rule of structural re-parameterization.
Therefore, the feature reuse process can be moved from feature space to weight space during inference, $i.e.$, to reuse features implicitly and efficiently. Based on RepGhost module, our hardware-efficient CNN \textit{RepGhostNet} outperforms state-of-the-art (SOTA) light-weight CNNs in accuracy-latency trade-off, as shown in Figure~\ref{fig:acc-to-latency}. Our contributions are summarized as:
% \vspace{-0.2cm}
\begin{itemize}
    \setlength\itemsep{-0.05em}
    \item We show that concatenation operation is not cost-free and indispensable for feature reuse in hardware-efficient architecture design, and propose a new perspective to realize feature reuse via structural re-parameterization technique.
    % \item We are the first to built hardware-efficient CNNs with re-parameterization technique, instead of its regular usage to boost performance~\cite{ding2019acnet,ding2021repvgg}, which does not change the network topology and latency.
    \item We are the first to utilize re-parameterization for simplifying network topology and improving hardware-efficiency, instead of its regular usage to boost performance~\cite{ding2019acnet,ding2021repvgg}, which does not change the network.
    \item We propose a novel RepGhost module with implicit feature reuse and develop a more hardware-efficient RepGhostNet compared to SOTA light-weight CNNs~\cite{howard2019searching,han2020ghostnet,ma2018shufflenet}. We show that RepGhostNet can achieve better accuracy-latency trade-off on several vision tasks.
    % , such as classification, object detection and instance segmentation.
    % \item We show that our RepGhostNet can achieve better performance on several vision tasks, such as classification, object detection and instance segmentation, compared to previous state-of-the-art light-weight CNNs with lower mobile latency.
\end{itemize}

%------------------------------------------------------------------------

\section{Related Work}
\label{sec:related-work}
\subsection{Light-weight CNNs}
Both manual designed~\cite{howard2017mobilenets,zhang2018shufflenet,sandler2018mobilenetv2} and neural architecture search (NAS)~\cite{wu2019fbnet,wan2020fbnetv2,guo2020single,hu2020angle} based light-weight CNNs are mainly designed to get competitive performances with less parameters and low FLOPs. Among them, ShuffleNetV1~\cite{zhang2018shufflenet} and MobileNetV2~\cite{sandler2018mobilenetv2} establish the benchmark by using massive depthwise convolutions rather than the dense ones. FBNet~\cite{wu2019fbnet,wan2020fbnetv2} employ complex NAS technique to design light-weight architecture automatically. However, parameters and FLOPs often can not reflect the actual runtime performance ($i.e.$, latency) of light-weight CNNs~\cite{ma2018shufflenet,dehghani2021efficiency}. Only few models are designed for low latency directly, like ProxylessNAS~\cite{kim2022proxyless}, MNASNet~\cite{tan2019mnasnet}, MobileNetV3~\cite{howard2019searching}, and ShuffleNetV2~\cite{ma2018shufflenet}. This paper follows this idea to design low-latency and efficient CNNs.

On the other hand, feature reuse in CNNs has also inspired many impressive works~\cite{huang2017densely,huang2018condensenet,han2020ghostnet,han2022ghostnets,ma2018shufflenet,szegedy2015going} with cheap or even free costs. As light-weight CNNs, GhostNet~\cite{han2020ghostnet} uses cheap operations to produce more channels with low computational costs, and ShuffleNetV2~\cite{ma2018shufflenet} processes only half of channels of features and keep the other half to be concatenated. They all use concatenation operation to keep large channel numbers since it is parameters- and FLOPs-free. But we note that it is inefficient on mobile devices due to its complicated memory copy process, making it not indispensable for feature reuse in light-weight CNNs. Therefore, in this paper, we explore to utilize feature reuse in light-weight CNNs architecture design beyond concatenation operation.

\subsection{Structural Re-parameterization}
Structural re-parameterization is generally to transform the more expressive and complex architecture at training time into a simpler one during inference, thus improving performance without any extra inference cost. ExpandNets~\cite{guo2020expandnets} expands the linear layers in the model into several continuous linear layers during training. ACNet~\cite{ding2019acnet} and RepVGG~\cite{ding2021repvgg} decompose a single convolutional layer into a training-time multi-branch block. For example, one such training-time block in RepVGG contains three parallel layers, $i.e.$, 3$\times$3 convolution, 1$\times$1 convolution and identity mapping, and an add operator to fuse their output features. During inference, the fusion process can be moved from feature space to weight space, resulting in a simpler block for fast inference (only one 3$\times$3 convolution)~\cite{ding2021repvgg}.
% This motivates us to design light-weight CNNs with efficient inference architectures using this technique.
Recently, this technique is also employed by MobileOne~\cite{vasu2023mobileone} to improve performance and design mobile backbones with large FLOPs for the powerful NPU in iPhone12.
% In this paper, however, we note that shortcut is necessary and will not bring much time-cost in extremely light-weight CNNs (See~\ref{subsec.ablation}). 

All of these works build~/~have their CNNs firstly, and then utilize structural re-parameterization technique to improve the performance, $e.g.$, for existing CNNs~\cite{ding2019acnet,guo2020expandnets} and specially designed CNNs with only 3$\times$3 convolution~\cite{ding2021repvgg} or without shortcuts~\cite{vasu2023mobileone}. However, instead of merely utilizing re-parameterization for performance gain, this paper explores to use this technique to reuse features implicitly and simplify network topology for fast inference.

\section{Method}
\label{sec:method} % key: feature reuse, rethinking re-parameterization, hardware-efficient

In this section, we will first revisit concatenation operation for feature reuse, and introduce how to utilize structural re-parameterization to achieve this. Based on it, we propose a novel re-parameterized module for implicit feature reuse, $i.e.$, RepGhost module. After that, we describe our hardware-efficient network built on this module, which is denoted as RepGhostNet. We also discuss the role of re-parameterization in our method, which is quite different from those of other works.

\subsection{Feature Reuse via Re-parameterization}
\label{subsec:feature-reuse}

Feature reuse has been widely used in CNNs to enlarge the network capacity, such as DenseNet~\cite{huang2017densely}, ShuffleNetV2~\cite{ma2018shufflenet} and GhostNet~\cite{han2020ghostnet}. Most methods utilize the concatenation operator combining feature maps from different layers to produce more features cheaply. Concatenation is parameters- and FLOPs-free, however, its computational cost is non-negligible due to the complicated memory copy on hardware devices. To address this, we provide a new perspective to realize feature reuse implicitly: \textit{feature reuse via structural re-parameterization}. 

\setcounter{footnote}{0}

\textbf{Concatenation costs.} As mentioned above, memory copy in concatenation brings non-negligible computational costs on hardware devices. For example, let $M_1 \in \mathbb{R}^{N \times C_1 \times H \times W}$ and $M_2 \in \mathbb{R}^{N \times C_2 \times H \times W}$ be two feature maps in data layout NCHW\footnote{Data layout NCHW4 is the same case as NCHW.} to be concatenated alone channel dimension. The largest contiguous blocks of memory required when processing $M_1$ and $M_2$ are $b_1 \in \mathbb{R}^{1 \times C_1 \times H \times W}$ and $b_2 \in \mathbb{R}^{1 \times C_2 \times H \times W}$, respectively. Concatenating $b_1$ and $b_2$ is direct, and this process would repeat $N$ times~\footnote{For batch size 1, pre-allocating memories can omit the copy process, but it needs operator-level optimization.}.
For data in layout NHWC~\cite{abadi2015tensorflow}, the largest contiguous block of memory is much smaller, i.e., $b_1 \in \mathbb{R}^{1\times 1 \times 1 \times {C_1}}$, making the copy process more complicated. While for element-wise operators, like Add, the largest contiguous blocks of memory are $M_1$ and $M_2$ themselves, making the operations much easier.

\begin{wrapfigure}{r}{0.5\columnwidth}	
	\vspace{-2mm}			
        \centering
	\small 
 	\captionof{table}{Runtime of concatenation and add operators with different batch sizes.} 
  	\vspace{-2mm}			
	\begin{tabular}{ccccc}
    \toprule
    \multirow{2}{*}{Operator} & \multicolumn{4}{c}{Latency(ms)} \\
    \cline{2-5}
     &  bs=1 & 2 & 8 & 32\\
    % \midrule
    \midrule
    Concatenation & 5.2 & 13.2 & 64.2 & 292.2\\
    Add & 2.6 & 6.7 & 34.0 & 149.8\\
    % -(ours) & 0 & 0 & 0 & 0\\
    \bottomrule
	\end{tabular}%
	\vspace{-2mm}
    \label{tab:concat-time}
\end{wrapfigure}

\begin{wrapfigure}{r}{0.48\columnwidth}	
	\vspace{-2mm}			
        \centering
	\small 
   \includegraphics[width=1\linewidth]{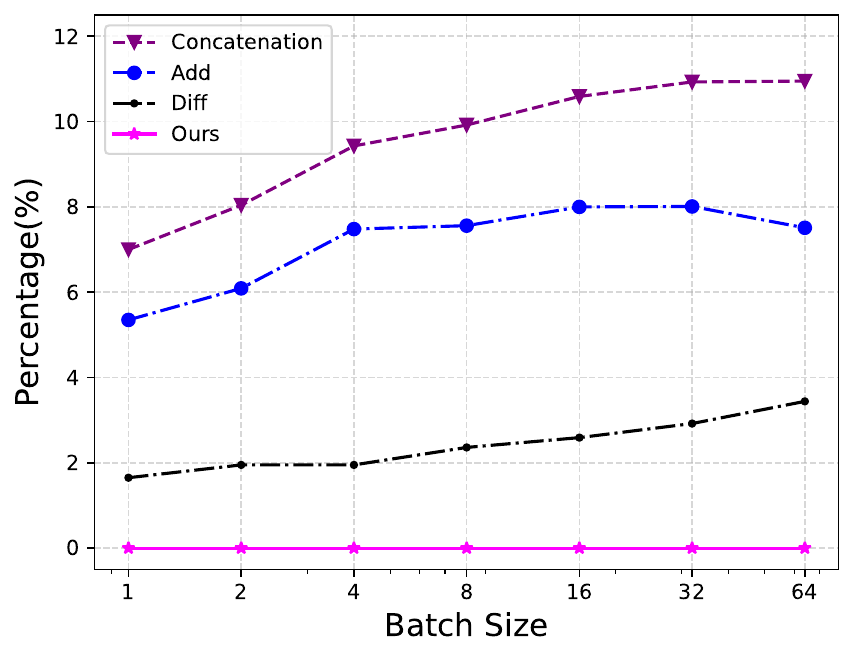} %{time-percentage-diff-0301.pdf}
	\vspace{-3mm}

  	\captionof{figure}{Runtime percentage of each operator in the entire network. \textbf{Diff:} the percent difference between concatenation and add. \textbf{Ours:} our method takes the add operator as an intermediate state, and it can be fused for fast inference.} 
	% \vspace{-3mm}
   \label{fig:time-percentage}
\end{wrapfigure}

To evaluate the concatenation operation quantitatively, we analyze its actual runtime on an ARM-based mobile device. We take GhostNet 1.0x~\cite{han2020ghostnet} as an example, and replace all its concatenation operators with add operators for comparison, which is also a simple operator to process different features with low costs~\cite{he2016deep,srivastava2015highway}.
Note that these two operators operate on tensors with exactly the same shape. Table~\ref{tab:concat-time} shows the accumulated runtime of all 32 corresponding operators in the corresponding network. Concatenation costs $\sim$2x times over Add. We also plot the time percentages under different batch sizes in Figure~\ref{fig:time-percentage}. With batch size increases, the gap of runtime percentage becomes larger, which is consistent with our data layout analysis.
% We hope this latency bottleneck at the concatenation operation could inspires more efficient CNNs architecture designs, \eg~, improving YOLO series models~\cite{wang2022yolov7,Jocher_YOLOv5_by_Ultralytics_2020,li2022yolov6}, which massively use concatenation for explicit feature reuse.

% \begin{figure}[t]
% % \vspace{-.25cm}
%   \centering
% %   \fbox{\rule{0pt}{2in} \rule{0.9\linewidth}{0pt}}
%    \includegraphics[width=0.6\linewidth]{time_percentage_diff_0301.pdf} %{time-percentage-diff-0301.pdf}
%    \vspace{-.25cm}
%    \caption{Runtime percentage of each operator in the entire network. \textbf{Diff:} the percent difference between concatenation and add. \textbf{Ours:} our method takes the add operator as an intermediate state, and it can be fused for fast inference.}
%    \label{fig:time-percentage}
%    % \vspace{-.25cm}
% \end{figure}

\textbf{Re-parameterization \textit{vs.} Concatenation} Let $y \in \mathbb{R}^{N \times C_{out} \times H \times W}$ denotes the output with $C_{out}$ channels and $x \in \mathbb{R}^{N \times C_{in} \times H \times W}$ the input to be processed and reused. $\Phi_i(x), \forall i=1, \ldots, s-1$ denote other layers, such as convolution or BN, applied to x. Without loss of generality, feature reuse via concatenation can be expressed as:

\vspace{-.15cm}
\begin{equation}
  % \vspace{-.1cm}
  y = Cat([x, \Phi_1(x), \ldots, \Phi_{s-1}(x)])
  % \vspace{-.15cm}
  \label{eq:feature-reuse-via-concat}
\end{equation}

where $Cat$ is the concatenation operation. It simply keeps existing feature maps and leaves the information processing to other operators. For example, a concatenation layer is usually followed by an 1$\times$1 dense convolutional layer to process the channel information~\cite{szegedy2015going,huang2017densely,han2020ghostnet}. However, as Table~\ref{tab:concat-time} shows, concatenation is not cost-free for feature reuse on hardware devices, which motivates us to find a more efficient way.

Recently, structural re-parameterization has been treated as a cost-free technique to improve the performance of CNNs in many works~\cite{ding2019acnet,ding2021repvgg,vasu2023mobileone}. Inspired by this, we note that structural re-parameterization can also be treated as an efficient technique for implicit feature reuse, so as to design more hardware-efficient CNNs. For example, structural re-parameterization usually utilizes several linear operators to produce diverse feature maps during training, and fuse all operators into one via parameters fusion for fast inference. That is, it moves the fusion process from feature space to weight space, which can be treated as an implicit way for feature reuse. Follow the symbols in Eq~\ref{eq:feature-reuse-via-concat}, feature reuse via structural re-parameterization can be expressed as:

% \vspace{-.3cm}
\begin{equation}
  y = Add([x, \Phi_1(x), \ldots, \Phi_{s-1}(x)]) = \Phi^{\ast}(x)
  \label{eq:feature-reuse-via-rep}
  % \vspace{-.15cm}
\end{equation}

Different from concatenation, add also plays a feature fusion role. All operation $\Phi_i(x), \forall i=1, \ldots, s-1$ in structural re-parameterization are linear function, and will be fused into $\Phi^{\ast}(x)$ finally. The feature fusion process is done in the weight space, which will not introduce any extra inference time, making the final architecture more efficient than that with concatenation or add operators.

As shown in Figure~\ref{fig:time-percentage}, our method implements feature reuse via structural re-parameterization. We not only discard the concatenation operator but also move the add process to weight space, thus saving $7\%\sim11\%$ time compared to concatenation and $5\%\sim8\%$ to add. Based on this concept, we propose a hardware-efficient module for feature reuse via re-parameterization in the next subsection.

\subsection{RepGhost Module}\label{subsection-RepGhost-module}
\begin{figure}[ht]
  \centering
  \vspace{-.45cm}
%   \fbox{\rule{0pt}{2in} \rule{0.9\linewidth}{0pt}}
   \includegraphics[trim=0cm 14cm 20.5cm 0cm, clip, width=0.7\linewidth]{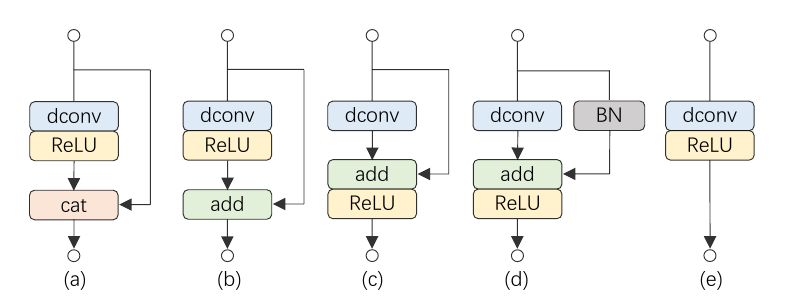}
   % \vspace{-.15cm}
   \caption{Evolution from Ghost module to RepGhost module. We omit the input 1$\times$1 convolution for simplicity, refer to Figure~\ref{fig:repghost-bottleneck} for more structure details. \textbf{dconv:} depthwise convolutional layer. \textbf{cat}: concatenation layer. a) Ghost module~\cite{han2020ghostnet} with ReLU; b) replacing concatenation with add; c) moving ReLU backward to make the module satisfying the rule of structural re-parameterization; d) \textbf{RepGhost module} during training; e) RepGhost module during inference.}
   % Module $c$ and $d$ can be fused into module $e$ during inference.}
   \label{fig:repghost-evolution}
   \vspace{-.15cm}
\end{figure}

To utilize feature reuse via re-parameterization, this subsection introduces how Ghost module evolves to our RepGhost module. It is non-trivial to apply re-parameterization to original Ghost module directly due to concatenation operator. As Figure~\ref{fig:repghost-evolution} shows, we start from Ghost module in Figure~\ref{fig:repghost-evolution}\textit{a} and several adjustments are made to derive our RepGhost module.

% adjust the intra components progressively.

\textbf{Add Operator.} Due to the inefficiency of concatenation for feature reuse discussed in Section~\ref{subsec:feature-reuse}, we first replace concatenation operator with add operator~\cite{he2016deep,srivastava2015highway} to get module $b$ in Figure~\ref{fig:repghost-evolution}. It should provide higher efficiency as shown in Table~\ref{tab:concat-time} and Figure~\ref{fig:time-percentage}.

\textbf{Moving ReLU Backward.} In the spirit of structural re-parameterization~\cite{ding2019acnet,ding2021repvgg}, we move the ReLU after depthwise convolutional layer backward, $i.e.$, after add operator, as module $c$ shown in Figure~\ref{fig:repghost-evolution}. This movement makes the module satisfying the rule of structural re-parameterization~\cite{ding2021repvgg,ding2019acnet}, and thus available to be fused into module $e$ for fast inference. We will discuss this in Section~\ref{subsec.ablation}.
% The effect of moving ReLU backward will be discussed in~\ref{subsec.ablation}.
% \footnote{A module is re-parameterizable, which we mean it can be fused into a module with a simplified topology.}

\textbf{Re-parameterization.} As a re-parameterized module, module $c$ can be more flexible in the re-parameterization structure rather than identity mapping~\cite{ding2019acnet,vasu2023mobileone}. As module $d$ shown in Figure~\ref{fig:repghost-evolution}, we simply add Batch Normalization(BN)~\cite{ioffe2015batch} in the identity branch, which brings non-linearity during training and can be fused during inference. It is denoted as our \textbf{RepGhost module}. We also explore other re-parameterization structures in Section~\ref{subsec.ablation}.

\textbf{Fast Inference.} As re-parameterized modules, module $c$ and module $d$ can be fused into module $e$ in Figure~\ref{fig:repghost-evolution} for fast inference. Our RepGhost module has a simple inference structure which only contains a regular convolutional layer and a ReLU, making it hardware-efficient~\cite{ding2021repvgg}. Specifically, the feature fusion process is carried out in weight space, $i.e.$, fusing parameter of each branch and producing a simplified topology for fast inference. Due to the linearity of each operator, the parameter fusion process is direct (see ~\cite{ding2019acnet,ding2021repvgg} for the detail).

\textbf{Comparison with Ghost module.} GhostNet~\cite{han2020ghostnet} proposes to generate more feature maps from cheap operations, thus enlarging the network capacity in a low-cost way. In our RepGhost module, we further propose a more efficient way to generate and fuse diverse feature maps via re-paramterization. Different from Ghost module, RepGhost module removes the inefficient concatenation operator, saving much inference time. And the information fusion process is executed by add operator in an implicit way, instead of leaving to other convolutional layers. 

Ghost module~\cite{han2020ghostnet} has a ratio $s$ to control the complexity. According to Eq~\ref{eq:feature-reuse-via-concat}, $C_{in}=\frac{1}{s}*C_{out}$ and the rest $\frac{s-1}{s}*C_{out}$ channels are produced by depthwise convolutions $\Phi_i, \forall i=1, \ldots, s-1$.
While for our RepGhost module, $C_{in}=C_{out}$.
% While the final outputs $C_{out}$ of our RepGhost are equal to $C_{in}$.
It produces diverse feature maps with $s*C_{in}$ channels during training same as Ghost module, but fuses them into $C_{in}$ channels for fast inference, and thus lower FLOPs.
% which not only keeps the generation of diverse feature maps, but also saves the inference time on hardware devices.
% To summary, we propose RepGhost module for feature reuse via structural re-parameterization while being efficient during inference, as stated in~\ref{eq:repghost-add}. As a basic module, RepGhost module is used to build our efficient CNNs, which will be introduced next.
That is, the difference between their output channels
% our RepGhost module differs to Ghost module mainly in their output channels, 
makes it non-trivial to improve Ghost module by directly using RepGhost module.
% We will introduce how to set channels properly to make full use of our RepGhost module in the next subsection.

\subsection{Building our Bottleneck and Architecture}\label{subsection-RepGhost-bottleneck-RepGhostnet}

\begin{figure}[t]
  \centering
%   \fbox{\rule{0pt}{2in} \rule{0.9\linewidth}{0pt}}
   \includegraphics[trim=0.1cm 6.1cm 18.05cm 0cm, clip, width=0.7\linewidth]{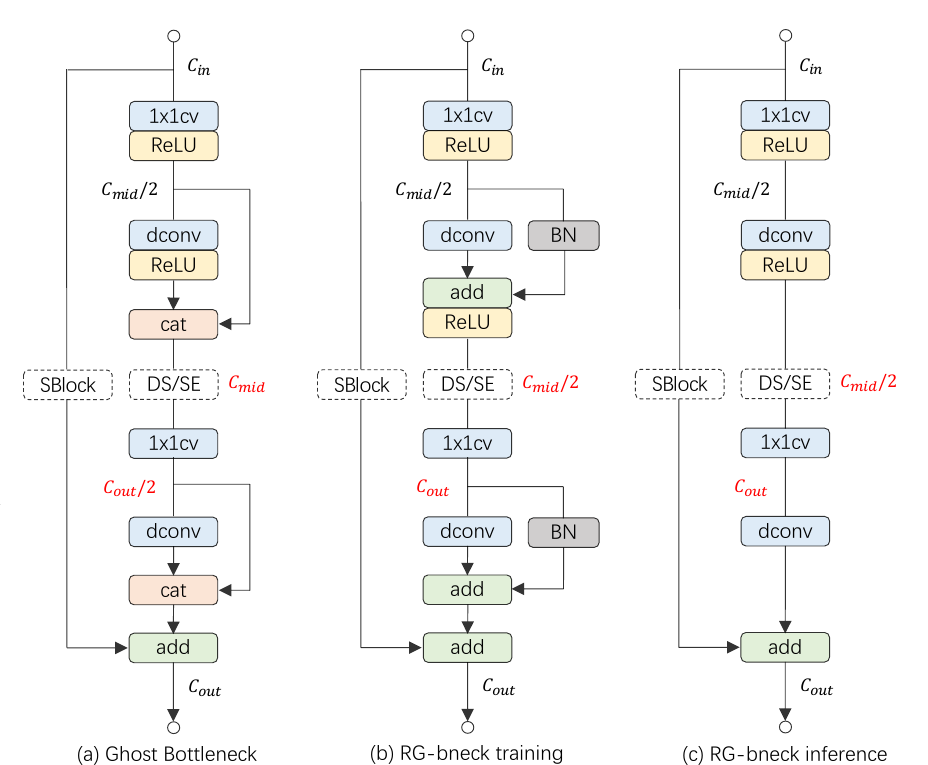}
   % \vspace{-.65cm}
   \caption{RepGhost bottleneck compared to Ghost bottleneck~\cite{han2020ghostnet}. \textbf{1$\times$1cv:} 1$\times$1 convolutional layer, \textbf{SBlock:} shortcut block, \textbf{DS:} downsample layer, \textbf{SE:} SE block~\cite{hu2018squeeze}. \textbf{RG-bneck:} RepGhost bottleneck. The blocks in dash line are only inserted if necessary. $C_{in}, C_{mid}$ and $C_{out}$ denote input, middle and output channels of the bottlenecks, respectively. Note that RepGhost bottleneck only differs to Ghost bottleneck in channels inside the bottlenecks, which are masked as {\color{red} red}.}
   % \vspace{-.1cm}
   \label{fig:repghost-bottleneck}
   % \vspace{-.5cm}
\end{figure}

\textbf{RepGhost Bottleneck.} Due to the change of output channels, this subsection introduces how to set channels properly to utilize RepGhost module to build our RepGhost bottleneck. As shown in Figure~\ref{fig:repghost-bottleneck}, our RepGhost bottleneck keeps the input and output channels of Ghost bottleneck~\cite{han2020ghostnet} and replaces the two Ghost modules with our RepGhost modules directly.
% , in which we keep the channel numbers of the first module, but double those of the second one.
% However, the add operator in RepGhost module differs to the concatenation operator in Ghost module in output channels, \eg~, 50\% of the latter. Simply changing the input channel of the following layer will change the network severely. To address this, 
% RepGhost bottleneck keeps the input and output channel numbers same as Ghost bottleneck.
As Figure~\ref{fig:repghost-bottleneck}\textit{b} shows, 
RepGhost bottleneck has two changes in channel setting: a) "thinner" middle channels, and b) "thicker" channels for second depthwise convolutional layer. Firstly, applying downsample and SE on feature maps with decreased channels makes RepGhost bottleneck more efficient~\cite{Radosavovic_2020_CVPR}. Secondly, applying  depthwise convolution on feature maps with increased channels enlarges the network capacity~\cite{sandler2018mobilenetv2}, thus making RepGhost bottleneck more effective.
During inference, RepGhost bottleneck only contains 2 branches (\ref{fig:repghost-bottleneck}\textit{c}): a shortcut and a single chain of operators (1$\times$1, depthwise convolution and ReLU), making it more efficient in memory cost and fast inference~\cite{ma2018shufflenet,ding2021repvgg}. We also extent the bottleneck to MobileNetV4~\cite{qin2024mobilenetv4} and verify its efficiency and effectiveness in Appendix~\ref{subsec:repghostv2}.
% We will profile the actual runtime on mobile devices to demonstrate the bottleneck's hardware efficiency.

\begin{table}[t]
  \centering
  \small
    \caption{Overall architecture of RepGhostNet. $\#$mid means the middle channel, it correspondences to $C_{mid}/2$ in Figure~\ref{fig:repghost-bottleneck}. $\#$out means the output channel. SE indicates whether to use SE blocks.}
  \begin{tabular}{cccccc}
    \toprule
    Input &  Operator & $\#$mid & $\#$out & SE & Stride\\
    % \midrule
    \midrule
    $224^2 \times 3$ & Conv$3\times3$ & - & 16 & - & 2\\
    \midrule
    $112^2 \times 16$ & RG-bneck & 8 & 16 & - & 1\\
    $112^2 \times 16$ & RG-bneck & 24 & 24 & - & 2\\
    \midrule
    $56^2 \times 24$ & RG-bneck & 36 & 24 & - & 1\\
    $56^2 \times 24$ & RG-bneck & 36 & 40 & 1 & 2\\
    \midrule
    $28^2 \times 40$ & RG-bneck & 60 & 40 & 1 & 1\\
    $28^2 \times 40$ & RG-bneck & 120 & 80 & - & 2\\
    \midrule
    $14^2 \times 80$ & RG-bneck & 100 & 80 & - & 1\\
    $14^2 \times 80$ & RG-bneck & 120 & 80 & - & 1\\
    $14^2 \times 80$ & RG-bneck & 120 & 80 & - & 1\\
    $14^2 \times 80$ & RG-bneck & 240 & 112 & 1 & 1\\
    $14^2 \times 112$ & RG-bneck & 336 & 112 & 1 & 1\\
    $14^2 \times 112$ & RG-bneck & 336 & 160 & 1 & 2\\
    \midrule
    $7^2 \times 160$ & RG-bneck & 480 & 160 & - & 1\\
    $7^2 \times 160$ & RG-bneck & 480 & 160 & 1 & 1\\
    $7^2 \times 160$ & RG-bneck & 480 & 160 & - & 1\\
    $7^2 \times 160$ & RG-bneck & 480 & 160 & 1 & 1\\
    $7^2 \times 160$ & Conv$1 \times 1$ & - & 960 & - & 1\\
    \midrule
    $7^2 \times 960$ & AvgPool & - & 960 & - & -\\
    \midrule
    $1^2 \times 960$ & Conv$1 \times 1$ & - & 1280 & - & 1\\
    \midrule
    $1^2 \times 1280$ & Conv$1 \times 1$ & - & 1000 & - & 1\\
    \bottomrule
  \end{tabular}
  % \vspace{-.25cm}

  \label{tab:repghostnet-arch}
  \vspace{-.5cm}
\end{table}

\textbf{RepGhostNet.} With the RepGhost bottleneck built above and its input and output channel numbers same as Ghost bottleneck, RepGhostNet can be simply built by replacing Ghost bottleneck in GhostNet~\cite{han2020ghostnet} with our RepGhost bottleneck. The architecture detail is shown in Table~\ref{tab:repghostnet-arch}. RepGhostNet stacks RepGhost bottlenecks except the input and output layers. A dense convolutional layer with 16 channels processes the input data, and a stack of normal 1$\times$1 convolutions and average pooling predicts the final outputs.
% We group RepGhost bottlenecks into 5 groups according to their inputs size, and stride=2 is set for last bottleneck in each group except for the last one. 
We slightly change the middle channels in group $4$ to keep channels non-decreasing in this group~\cite{Radosavovic_2020_CVPR}.
We also apply SE block~\cite{hu2018squeeze} and use ReLU as the non-linearity function in RepGhostNet as GhostNet~\cite{han2020ghostnet} does. Following ~\cite{ma2018shufflenet,han2020ghostnet}, a \textit{width multiplier} $\alpha$ is used to scale the network, which is denoted as RepGhostNet $\alpha\times$.

\subsection{Re-parameterization for Fast Inference}

Our RepGhostNet is built with re-parameterization for implicit feature reuse, making its topology to be able to be further simplified for fast inference. However, re-parameterization is often used to improve performances and does not change their topology and latency~\cite{ding2019acnet,vasu2023mobileone}. But we note that this performance gains on light-weight CNNs are marginal, as shown in Table~\ref{tab:reparam-mbv3-ghost}, in which we apply re-parameterization to MobileNetV3~\cite{howard2019searching} and GhostNet~\cite{han2020ghostnet} same as RepGhostNet. It is interesting that re-parameterization brings no performance gain to GhostNet, which is designed to reuse features explicitly via concatenation.
While re-parameterization also reuses features, but implicitly, so it does not benefit GhostNet more.

\begin{wrapfigure}{r}{0.55\columnwidth}	
	\vspace{-4mm}	
        \centering
	\small 
 	\captionof{table}{Effects of re-parameterization on two light-weight CNNs.} 
  \vspace{-2mm}
  \begin{tabular}{ccl}  % l}
    \toprule
    Model & Reparam & Top1 Acc.(\%) \\% & training (h) \\ % & $\Delta$\\
    % \midrule
    % \midrule
    % \multirow{2}{*}{RGhost 0.58$\times$}& \ding{55} &  \\
    % & \ding{51} & 68.9(+)\\
    \midrule
    % \shline
    \multirow{2}{*}{MobileNetV3 Small 1.0$\times$}& \ding{55} & 67.7\\
    & \ding{51} & 68.0(+0.3) \\
    \midrule
    % \shline
    \multirow{2}{*}{GhostNet 0.5$\times$}& \ding{55} & 66.4\\
    & \ding{51} & 66.4(+0.0) \\
    \bottomrule
  \end{tabular}
  \vspace{-.35cm}

  \label{tab:reparam-mbv3-ghost}
\end{wrapfigure}

Our RepGhostNet, however, is the first to reuse features implicitly via re-parameterization technique and produces a simplified topology for fast inference. As we will show in Table~\ref{tab:main-results}, compared to GhostNet 0.5$\times$, 1.0$\times$ and 1.3$\times$, our RepGhostNet 0.5$\times$, 1.0$\times$ and 1.3$\times$ get not only 0.2$\sim$0.5\% higher accuracy but also significant speedup, $i.e.$, \textbf{16.5$\sim$21.0\%} faster on the mobile device.

%------------------------------------------------------------------------
\section{Experiments}
\label{sec:exp}

In this section, to show the superiority of the proposed RepGhostNet, we evaluate the architecture on ImageNet 2012 classification benchmark~\cite{deng2009imagenet}, MS COCO 2017 object detection and instance segmentation benchmarks~\cite{lin2014microsoft} and make a fair comparison with other SOTA light-weight CNNs.
% We evaluate the latency of all models on an ARM-based mobile device using tensor compute engine MNN~\cite{alibaba2020mnn}.
% Specifically, the device is Xiaomi 5$\times$ with Qualcomm Snapdragon 625 processor.

\textbf{Datasets.} ImageNet has been a standard benchmark for visual models. It contains 1,000 classes with 1.28M training images and 50k validation images. We use all the training data and evaluate models on the validation images. Top-1 and Top-5 accuracy with single crop are reported.

MS COCO is also a well-known visual benchmark. We train our models using COCO 2017 $trainval35k$ split and evaluate on the $minival$ split with 5K images, following the open-source mmdetection~\cite{chen2019mmdetection} library.

\textbf{Latency.} Tensor compute engine MNN~\cite{alibaba2020mnn} is a light-weight framework for deep learning and provides efficient inference on mobile devices. So we use MNN to evaluate the latency of the models on the ARM-based mobile device using single thread. Batch size is set to 1 by default if not stated. Each model runs for 100 times and the latency is recorded as their average. Specifically, the used mobile device is \textbf{Xiaomi 5$\times$ with Qualcomm Snapdragon 625 processor}.
More latency evaluations on other mobile devices will be provided in the Appendix~\ref{sec:more-latency-eval}, including 3 more Android devices with different computational resources and an iPhone12.

\begin{table*}[t]
  \centering
  \small
  \caption{Classification results on ImageNet. We compare RepGhostNet to SOTA light-weight CNNs.}
    \vspace{-.2cm}
  \begin{tabular}{cccccc}
    \toprule
    Model &  Params(M) & FLOPs(M) & Latency(ms) & Top-1 (\%) & Top-5 (\%)\\
    \midrule
    % \midrule
    % MobileNetV2 0.35$\times$\cite{sandler2018mobilenetv2} & 1.7 & 59 & 36.0 & 60.3 & 82.9\\
    ShuffleNetV2 0.5$\times$\cite{ma2018shufflenet} & 1.4 & 41 & - & 61.1 & 82.6\\
    MobileNetV2 0.5$\times$\cite{sandler2018mobilenetv2} & 2.0 & 97 & 41.9 & 65.6 & 86.3\\
    MobileNeXt 0.35$\times$\cite{zhou2020rethinking} & 1.8 & 80 & 34.1 & 67.7 & -\\
    MobileNetV3 Small 0.75$\times$\cite{howard2019searching} & 2.0 & 44 & 26.0 & 65.4 & 86.0\\
    MobileNetV3 Small 1.0$\times$\cite{howard2019searching} & 2.5 & 57 & 31.9 & 67.7 & 87.5\\
    GhostNet 0.5x\cite{han2020ghostnet} & 2.6 & 42 & 31.7 & 66.4 & 86.6\\
    % FBNetV2-F1\cite{wan2020fbnetv2} & 6.0 & 56 & 32.2 & 68.3 & -\\
    \textbf{RepGhostNet 0.5$\times$} & 2.3 & 43 & \textbf{25.1} & \textbf{66.9} & \textbf{86.9}\\
    \textbf{RepGhostNet 0.58$\times$} & 2.5 & 60 & \textbf{31.9} & \textbf{68.9} & \textbf{88.4}\\

    \midrule
    % MobileNetV1 0.5$\times$\cite{howard2017mobilenets} & 1.3 & 150 & - & 63.3 & 84.9\\
    % MobileNetV2 0.6$\times$\cite{sandler2018mobilenetv2} & 2.2 & 141 & 76.7 & 66.7 & -\\
    ShuffleNetV2 1.0$\times$\cite{ma2018shufflenet} & 2.6 & 146 & - & 69.4 & 88.9\\
    MobileNetV2 0.75$\times$\cite{sandler2018mobilenetv2} & 2.2 & 141 & 78.2 & 70.6 & 89.6\\
    % FBNetV2-F3\cite{wan2020fbnetv2} & 6.9 & 126 & 61.2 & 73.2 & -\\
    MobileNetV3 Large 0.75$\times$\cite{howard2019searching} & 4.0 & 155 & 72.0 & 73.5 & 91.2 \\
    GhostNet 1.0$\times$\cite{han2020ghostnet} & 5.2 & 141 & 74.5 & 74.0 & 91.5\\
    \textbf{RepGhostNet 1.0$\times$} & 4.1 & 142 & \textbf{62.2} & \textbf{74.2} & \textbf{91.5}\\
    \textbf{RepGhostNet 1.11$\times$} & 4.5 & 170 & \textbf{71.5} & \textbf{75.1} & \textbf{92.2}\\
    \midrule
    ShuffleNetV2 1.5$\times$\cite{ma2018shufflenet} & 3.5 & 299 & - & 72.6 & 90.6\\
    MobileNetV2 1.0$\times$\cite{sandler2018mobilenetv2} & 3.5 & 300 & 106.7 & 73.3 & 91.1\\
    % MobileNeXt 0.75$\times$\cite{zhou2020rethinking} & 2.5 & 210 & 80.1 & 72.0 & -\\
    MobileNeXt 1.0$\times$\cite{zhou2020rethinking} & 3.4 & 300 & 113.6 & 74.0 & -\\
    MobileOne-S0\cite{vasu2023mobileone} & 2.1 & 275 & 79.4 & 71.4 & -\\
    % MobileNeXt plus 1.0$\times$\cite{zhou2020rethinking} & 3.9 & 330 & - & 76.1 & -\\
    ProxylessNAS\cite{kim2022proxyless} & 4.1 & 320 & 128.0 & 74.6 & 92.2\\
    EfficientNet B0\cite{tan2019efficientnet} & 5.3 & 386 & 298.0 & 77.1 & 93.3\\
    MobileNetV3 Large 1.0$\times$\cite{howard2019searching} & 5.4 & 219 & 95.2 & 75.2 & 92.4 \\
    % FBNetV2-F4\cite{wan2020fbnetv2} & 7.0 & 238 & 144.5 & 76.0 & -\\
    % FBNetV2-L1\cite{wan2020fbnetv2} & 8.5 & 325 & 190.5 & 77.2 & -\\
    GhostNet 1.3$\times$\cite{han2020ghostnet} & 7.3 & 226 & 117.6 & 76.2 & 92.9\\
    \textbf{RepGhostNet 1.3$\times$} & 5.5 & 231 & \textbf{92.9} & \textbf{76.4} & \textbf{92.9}\\
    \textbf{RepGhostNet 1.5$\times$} & 6.6 & 301 & \textbf{116.9} & \textbf{77.5} & \textbf{93.5}\\

    \bottomrule
  \end{tabular}
  % \vspace{-.35cm}
%   We compare RepGhostNet to state-of-the-art light-weight CNNs in terms of the number of parameters, FLOPs, latency, and accuracy.
  \label{tab:main-results}
  \vspace{-.45cm}
\end{table*}

\subsection{ImageNet Classification}
To demonstrate the effectiveness and efficiency of our proposed RepGhostNet, we compare to SOTA light-weight CNNs in terms of accuracy on ImageNet benchmark and latency on mobile devices.
% The competitors include MobileNet series~\cite{sandler2018mobilenetv2,howard2019searching}, ShuffleNetV2~\cite{ma2018shufflenet}, GhostNet~\cite{han2020ghostnet}, MobileNeXt~\cite{zhou2020rethinking}, MobileOne~\cite{vasu2023mobileone}, ProxylessNAS~\cite{kim2022proxyless}, and EfficientNet~\cite{tan2019efficientnet}.
% The training detail is provided below.

\textbf{Implementation Details.} We adopt PyTorch~\cite{paszke2019pytorch} and timm~\cite{rw2019timm} library for training. The global batch size is set to 1024 on 8 NVIDIA V100 GPUs. Standard SGD with momentum coefficient of 0.9 is the optimizer. Base learning rate is 0.6 and cosine anneals for 300 epochs with first 5 epochs for warming up, and weight decay is set as 1e-5. Dropout rate before classifier layer is set to 0.2.
We also use EMA (Exponential Moving Average) weight averaging with 0.9999 factor.
% TODO: data aug
For data augmentation, except regular image crop and flip in timm, we also utilize random erase with prob 0.2. For larger models, $e.g.$, RepGhostNet 1.3$\times$ (231M), auto-augmentation~\cite{cubuk2018autoaugment} is applied. For fair comparison, we also retrain MobileNetV2, MobileNetV3 and GhostNet using our training settings.

% All models are grouped into three levels according to FLOPs~\cite{han2020ghostnet}. The corresponding latency of each model is evaluated on Xiaomi 5$\times$. \ref{fig:acc_to_latency} plots the latency and accuracy of all models. We can see that our RepGhostNet outperforms other manually designed and NAS-based models in terms of accuracy-latency trade-off.

\textbf{Effective and Efficient.} As Figure~\ref{fig:acc-to-latency} and Table~\ref{tab:main-results} show, RepGhostNet outperforms
other SOTA light-weight CNNs in terms of accuracy-latency trade-off, including manually designed and NAS-based ones, $e.g.$, RepGhostNet 0.5$\times$ is \textbf{20\%} faster than GhostNet 0.5$\times$ with 0.5\% higher Top-1 accuracy, and RepGhostNet 1.0$\times$ is 14\% faster than MobileNetV3 Large 0.75$\times$ with 0.7\% higher Top-1 accuracy. With comparable latency, RepGhostNet surpasses all models by a large margin in all FLOPs levels, $e.g.$, our RepGhostNet 0.58$\times$ surpasses GhostNet 0.5$\times$ by \textbf{2.5\%} Top-1 accuracy.
% and RepGhostNet 1.11$\times$ surpasses MobileNetV3 Large 0.75$\times$ by 1.6\% Top-1 accuracy.

\subsection{Object Detection and Instance Segmentation}

To verify the generalization of our RepGhostNet as a general backbone, we conduct experiments on COCO~\cite{lin2014microsoft} object detection and instance segmentation benchmarks using mmdetection~\cite{chen2019mmdetection} library and compare with several other backbones in the tasks.

\textbf{Implementation Details.}
We use YOLOv3~\cite{redmon2018yolov3} and RetinaNet~\cite{lin2017focal}, and Mask RCNN~\cite{he2017mask} baselines for detection task and instance segmentation task, respectively. 
Following~\cite{chen2019mmdetection}, we only replace the ImageNet-pretrained backbones and train the models for 12 epochs in 8 NVIDIA V100 GPUs. Synchronized BN is also enabled. We report the mAP@IoU of 0.5:0.05:0.95 and evaluate the latency of single-stage models, $i.e.$, YOLOv3 and RetinaNet.
% We also report the latency (follow~\ref{tab:main-results}) of each backbone for reference.

\textbf{Results.}
As the results shown in \ref{tab:coco-results}, our RepGhostNet outperforms MobileNetV2~\cite{sandler2018mobilenetv2}, MobileNetV3~\cite{howard2019searching}, and GhostNet~\cite{han2020ghostnet} in both tasks in terms of accuracy-latency trade-off. For example, with comparable latency, RepGhostNet 1.3$\times$ surpasses all other backbones in both tasks clearly, and RepGhostNet 1.1$\times$ achieves comparable or even better performance with significant speedup.
% RepGhostNet 1.3$\times$ surpasses GhostNet 1.1$\times$ by more than 1\% mAP in both tasks and RepGhostNet 1.5$\times$ surpasses MobileNetV2 1.0$\times$ by more than 2\% mAP in both tasks.

\vspace{-.15cm}
\begin{table*}[t]
  \centering
  \small
  \caption{Detection and instance segmentation results on COCO dataset.}
  % YOLOv3 and RetinaNet are with size of 416 and 1024, respectively.}
   % We omit that of the two-stage method Mask RCNN for simplicity.
     \vspace{-.15cm}
  \begin{tabular}{cccccccccc}
    \toprule
    \multirow{2}{*}{Backbone} & & \multicolumn{2}{c}{YOLOv3} & & \multicolumn{2}{c}{RetinaNet} & & \multicolumn{2}{c}{Mask RCNN} \\
    \cline{9-10}
    \cline{6-7}
    \cline{3-4}
    & & Latency(s) & $mAP$ & & Latency(s) &  $mAP$ & & $mAP^{bbox}$& $mAP^{mask}$\\
    % \midrule
    \midrule
    MobileNetV2 1.0$\times$  & & 585.5 & 23.9 & & 3.85  & 32.1  & & 34.2 & 31.3\\
    MobileNetV3 1.0$\times$  & & 580.3 & 23.8 & & 3.70  & 32.7 & & 34.2 & 31.5\\
    GhostNet 1.1$\times$  & & 559.0 & 23.9 & & 3.70  & 32.5 & & 34.8 & 31.9\\
    RepGhostNet 1.11$\times$  & & \textbf{497.1} & \textbf{23.9} & & \textbf{3.63} & \textbf{32.7}  & & \textbf{35.0} & \textbf{32.0}\\
    RepGhostNet 1.3$\times$  & & \textbf{555.3} & \textbf{24.7} & & \textbf{3.64} & \textbf{33.5} & & \textbf{36.1} & \textbf{33.1}\\
    % RepGhostNet 1.5$\times$ & \textbf{116.9} & & 637.4 & 24.9 & & \textbf{3.78} & \textbf{34.4} & & \textbf{36.9} & \textbf{33.5}\\
    \bottomrule
  \end{tabular}
  \vspace{-.35cm}

  % Latency of the two-staged method Mask RCNN are not reported.
  \label{tab:coco-results}
  % \vspace{-.25cm}
\end{table*}

\subsection{Ablation Study}
\label{subsec.ablation}

% In this subsection, we verify our architecture design.
% Firstly, we compare RepGhost module and Ghost module in terms of the generalization to large model ResNet50~\cite{he2016deep}. 
% Firstly, we compare RepGhostNet to MobileOne~\cite{vasu2023mobileone}, a recent mobile CNN designed for iPhone, in terms of accuracy and latency on iPhone12.
% We then test and verify different re-parameterization structures of RepGhost module. Lastly, considering shortcut (or skip connection) is being gradually discarded by modern CNNs~\cite{ding2021repvgg,vasu2023mobileone}, we discuss a question: \textit{is shortcut necessary for light-weight CNNs?}.
% which is concerned in CNNs architecture designs recently.

\begin{wrapfigure}{r}{0.55\columnwidth}	
	\vspace{-2mm}			
        \centering
	\small 
 	\captionof{table}{Accuracy and latency on iPhone12.} 
  \vspace{-2mm}
  \begin{tabular}{cccl}  % l}
    \toprule
    % \multirow{2}{*}{Acc.(\%)}
    \multirow{2}{*}{Model} & Top1  &  \multicolumn{2}{c}{Latency(ms)} \\ % & $\Delta$\\
    \cline{3-4}
    & Acc.(\%) & NPU & CPU \\
    
    % \midrule
    \midrule
    MobileOne-S2 & 77.4 & 1.20 & 12.18\\
    RepGhostNet 1.5$\times$ & \textbf{77.5} & \textbf{1.15} & \textbf{9.15(+25\%)}\\
    \midrule
    MobileOne-S3 & 78.1 & 1.50 & 15.97\\
    RepGhostNet 2.0$\times$ & \textbf{78.8} & \textbf{1.48} & \textbf{10.90(+32\%)}\\
    \bottomrule
  \end{tabular}
	\vspace{-2mm}
  \label{tab:iphone-latency}
\end{wrapfigure}

\textbf{Comparison to MobileOne.} We evaluate RepGhostNet on an iPhone12 to compare with MobileOne. As shown in Table~\ref{tab:iphone-latency}, RepGhostNet 1.5$\times$ and 2.0$\times$ outperforms MobileOne in accuracy-latency trade-off, especially on iPhone12 CPU. Besides, we note that the reported latency in MobileOne paper~\cite{vasu2023mobileone} are evaluated on NPU, while we use CPU, which we conjecture causes the latency difference. More detailed comparisons are provided in Appendix~\ref{subsec:appendix-mobileone}.

\begin{wrapfigure}{r}{0.5\columnwidth}	
	\vspace{-2mm}			
        \centering
	\small 
 	\captionof{table}{Results of different re-parameterization structures on RepGhostNet 0.5$\times$.} 
  \vspace{-2mm}
        \label{tab:reparam-settings}
  \begin{tabular}{c|ccc|c}
    \toprule
    Variant & id & 1$\times$1dconv & BN & Top-1(\%)\\
    % \midrule
    \midrule
    no-reparam&&&&66.3\\
    \midrule
    &\ding{51}&&&66.4\\
    &&\ding{51}&&66.2\\
    &&&\ding{51}&\textbf{66.9}\\
    &&\ding{51}&\ding{51}&66.5\\
    % &\ding{51}&&\ding{51}&-\\
    &\ding{51}&\ding{51}&\ding{51}&66.5\\
    \midrule
    \color[gray]{0.6}{+ReLU}& \color[gray]{0.6}{} & \color[gray]{0.6}{} &\color[gray]{0.6}{\ding{51}}&\color[gray]{0.6}{66.6}\\
    \bottomrule
  \end{tabular}
    \vspace{-2mm}
\end{wrapfigure}

\textbf{Re-parameterization Structures.} To verify the re-parameterization structures of RepGhostNet, we alternate the components in the identity mapping branch (of module in Figure~\ref{fig:repghost-evolution}\textit{c}) of RepGhostNet 0.5$\times$, such as BN, 1$\times$1 depthwise convolution and identity mapping itself~\cite{ding2021repvgg,vasu2023mobileone,ding2019acnet}.
Table~\ref{tab:reparam-settings} shows that re-parameterization with BN achieves the best performance, and we take it as our default structure for all other RepGhostNets.
We attribute this improvement to the training-time non-linearity of BN, which provides more information than identity mapping.
The 1$\times$1 depthwise convolution is also followed by BN, therefore, its parameters have no effort on the features due to the following normalization and may make the BN statistics unstable, which may result in its poor performance.
%the poor performance may be caused by their coupling during optimization.
Note that all the models in Table~\ref{tab:reparam-settings} can be fused into a same efficient inference model, expect the last one row. Specially, we insert a ReLU after 3$\times$3 convolution like Ghost module to verify that it is safe to move the ReLU backward, as we did in Figure~\ref{fig:repghost-evolution}\textit{c}.
Besides, compared to MobileOne~\cite{vasu2023mobileone}, our re-parameterization structure is much simpler, $i.e.$, only one BN layer. The simpler re-parameterization structure makes negligible the additional training costs. For example, training RepGhostNet 0.5$\times$ with and without re-parameterization for 300 epochs cost 25.0 and 24.8 hours, respectively.

\section{Conclusion}
To utilize feature reuse in light-weight CNNs architecture design efficiently, this paper proposes a new perspective to realize feature reuse implicitly via structural re-parameterization technique, instead of the widely-used but inefficient concatenation operation. With this technique, a novel and hardware-efficient RepGhost module for implicit feature reuse is proposed. The proposed RepGhost module fuses features from different layers at training time, and carry out the fusion process in the weight space before inference, resulting in a simplified and hardware-efficient architecture for fast inference. Built on RepGhost module, we develop a hardware-efficient light-weight CNNs named RepGhostNet, which achieves new SOTA on several vision tasks in terms of accuracy-latency trade-off for mobile devices.

%------------------------------------------------------------------------

% \section*{References}

% References follow the acknowledgments in the camera-ready paper. Use unnumbered first-level heading for
% the references. Any choice of citation style is acceptable as long as you are
% consistent. It is permissible to reduce the font size to \verb+small+ (9 point)
% when listing the references.
% Note that the Reference section does not count towards the page limit.
% \medskip

% {
% \small

% [1] Alexander, J.A.\ \& Mozer, M.C.\ (1995) Template-based algorithms for
% connectionist rule extraction. In G.\ Tesauro, D.S.\ Touretzky and T.K.\ Leen
% (eds.), {\it Advances in Neural Information Processing Systems 7},
% pp.\ 609--616. Cambridge, MA: MIT Press.

% [2] Bower, J.M.\ \& Beeman, D.\ (1995) {\it The Book of GENESIS: Exploring
%   Realistic Neural Models with the GEneral NEural SImulation System.}  New York:
% TELOS/Springer--Verlag.

% [3] Hasselmo, M.E., Schnell, E.\ \& Barkai, E.\ (1995) Dynamics of learning and
% recall at excitatory recurrent synapses and cholinergic modulation in rat
% hippocampal region CA3. {\it Journal of Neuroscience} {\bf 15}(7):5249-5262.
% }

% \bibliography{repghost}
% \bibliographystyle{neurips_2024}

{\small 
	\bibliography{repghost}
 	\bibliographystyle{plain}
}
%%%%%%%%%%%%%%%%%%%%%%%%%%%%%%%%%%%%%%%%%%%%%%%%%%%%%%%%%%%%
\newpage

\appendix
\section*{Appendix}

%------------------------------------------------------------------------
\section{More Latency Evaluations}
\label{sec:more-latency-eval}
% \textbf{More mobile phones.}
\subsection{More mobile phones}
We evaluate the latency of all models in main paper on Xiaomi 5$\times$ with \textbf{Qualcomm Snapdragon 625 processor}, which is considered as a low-end one nowadays. To verify the generalization of our RepGhost module and RepGhostNet to other mobile devices, we evaluate the latency on other three Android mobile phones with more powerful processors. They are Xiaomi Note3, Xiaomi 8 and Huawei P20 with \textbf{Qualcomm Snapdragon 660 processor}, \textbf{Qualcomm Snapdragon 845 processor} and \textbf{Kirin 970 processor}, respectively.

\begin{figure*}[ht]
    \centering
    \begin{subfigure}[t]{0.45\textwidth}
    \centering
   \includegraphics[trim=0cm 0cm 0cm 0cm, clip, width=1.0\linewidth]{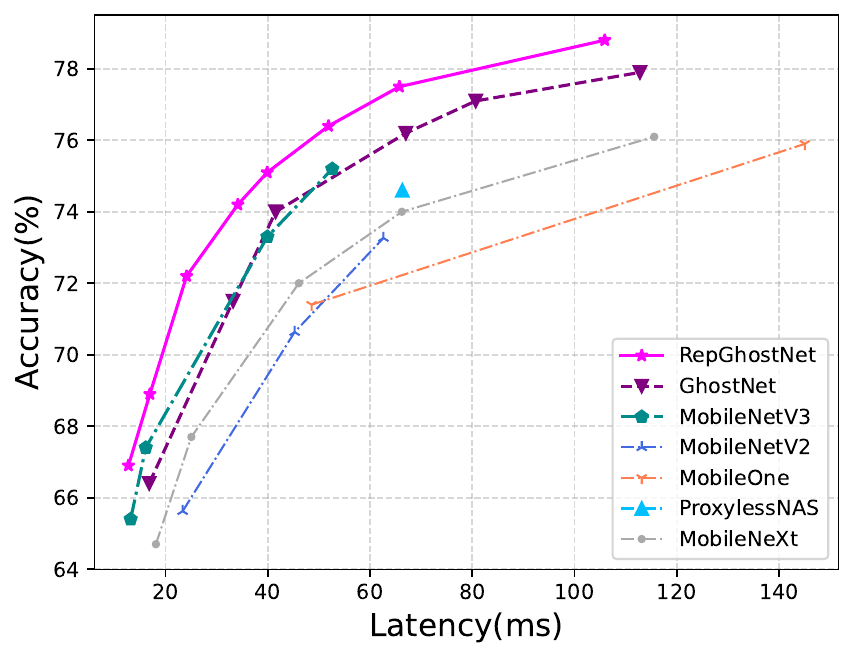}
   \caption{Xiaomi Note3 with the \textbf{Snapdragon 660} processor.}
  \label{fig:acc-to-latency-660}
    \end{subfigure}%
    \hfill 
    \begin{subfigure}[t]{0.45\textwidth}
        \centering
   \includegraphics[trim=0cm 0cm 0cm 0cm, clip, width=1.0\linewidth]{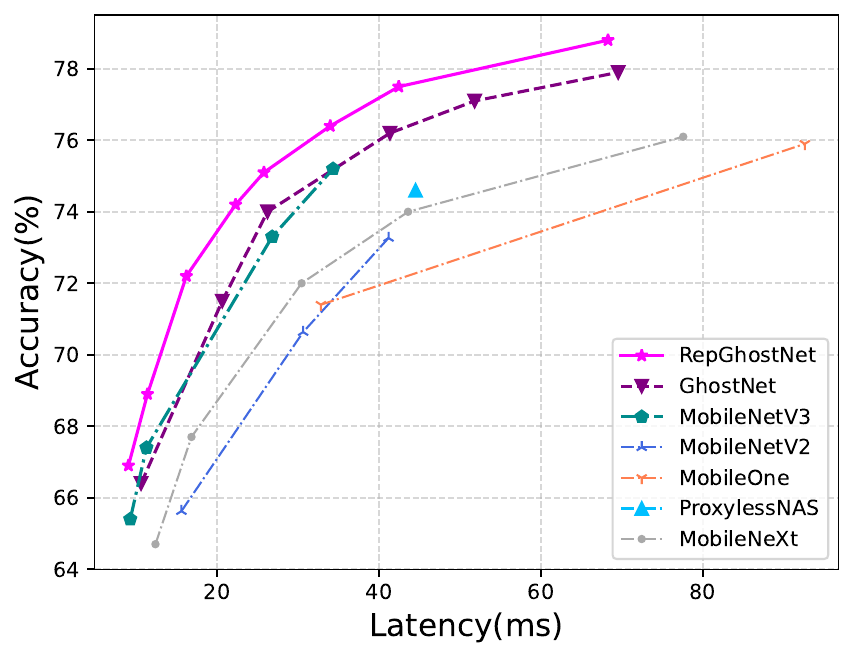}
   \caption{Xiaomi 8 with the \textbf{Snapdragon 845} processor.}
      \label{fig:acc-to-latency-845}
    \end{subfigure}
    \hfill 
    \begin{subfigure}[t]{0.45\textwidth}
        \centering
   \includegraphics[trim=0cm 0cm 0cm 0cm, clip, width=1.0\linewidth]{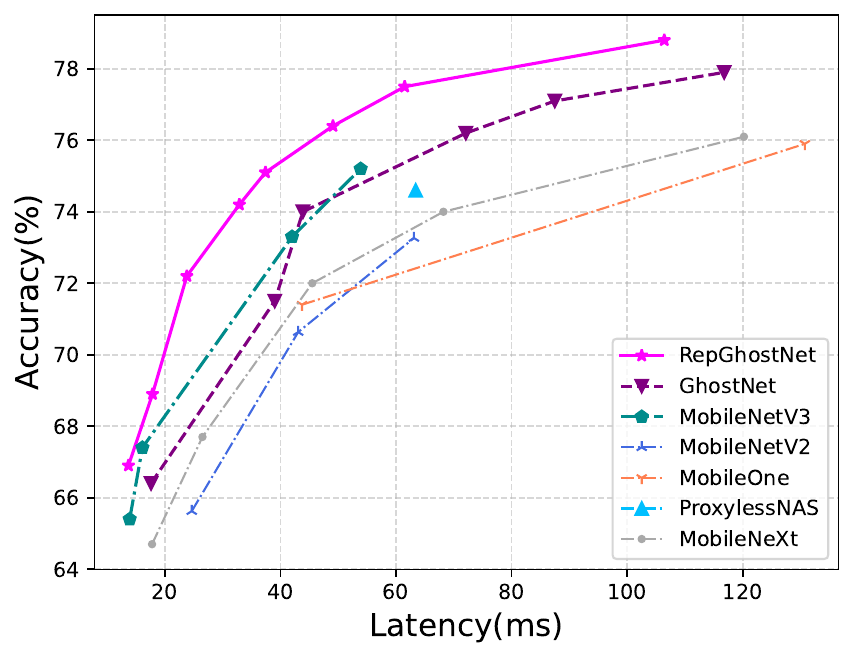}
   \caption{Huawei P20 with the \textbf{Kirin 970} processor.}
      \label{fig:acc-to-latency-970}
    \end{subfigure}
    \hfill 
    \begin{subfigure}[t]{0.45\linewidth}
        \centering
   \includegraphics[trim=0cm 0cm 0cm 0cm, clip, width=1.0\linewidth]{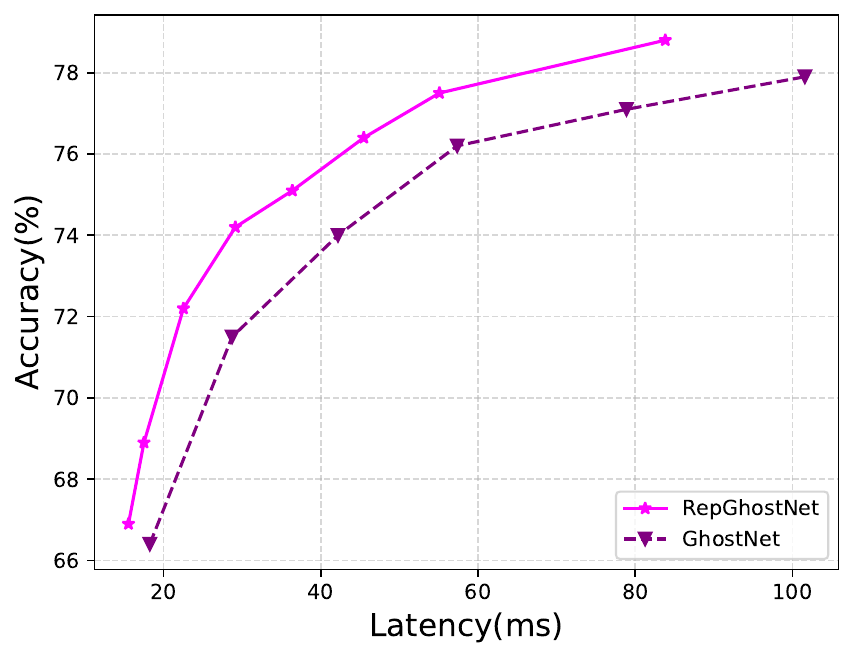}
   \caption{\textbf{TFLite}.}
\label{fig:acc-to-latency-repghost-ghost-kirin970-tflite}
    \end{subfigure}
    \caption{Top-1 accuracy on ImageNet \textit{vs.} latency of different devices (a,b,c) or compute engine (d).}
\end{figure*}

We plot the accuracy-latency results in Figure~\ref{fig:acc-to-latency-660}, Figure~\ref{fig:acc-to-latency-845} and Figure~\ref{fig:acc-to-latency-970} same as Figure~\ref{fig:acc-to-latency} in main paper.
% We also show it in Fig. \ref{fig:acc_to_latency}.
From these results, we can see similar trends of the curves, $i.e.$, our RepGhostNet outperforms other state-of-the-art light-weight CNNs in accuracy-latency trade-off in all four level mobile phones.
Besides, note that the results vary widely between the four mobile phones, $e.g.$, the latency of RepGhostNet 1.0$\times$ on them vary from 22.3ms to 62.2ms, which means that our RepGhost module and RepGhostNet generalize well to wide range of mobile devices with different computational resources.

% \textbf{Evaluation with TFLite.}
\subsection{Evaluation with TFLite}
\label{subsec:tflie}
While we use MNN~\cite{alibaba2020mnn} as our mobile compute engine for all models, we also evaluate the models using TFLite\footnote{\url{https://github.com/tensorflow/tensorflow/tree/master/tensorflow/lite/tools/benchmark/android}}~\cite{abadi2015tensorflow} in Figure~\ref{fig:acc-to-latency-repghost-ghost-kirin970-tflite}. We only compare RepGhostNet to GhostNet for convenience. As Figure~\ref{fig:acc-to-latency-repghost-ghost-kirin970-tflite} shows, we can observer similar trend of accuracy-latency as using MNN, verifying the generalization of our method. For example, compared to GhostNet 1.0\(\times\) and 1.5\(\times\), RepGhostNet 1.0\(\times\) and 1.5\(\times\) gets not only 0.2\% and 0.4\% higher Top1 accuracy, but also \textbf{31.0\% and 30.2\% speedup}, respectively. Compared to MNN, TFLite brings  more speedup of our RepGhostNet over GhostNet, thanks to the data layout of NHWC in TFLite, which makes the concatenation process in GhostNet more complicated, as we stated in Section~\ref{subsec:feature-reuse} in main paper. Therefore, the elimination of the latency bottleneck at concatenation operation in our RepGhostNet brings much better latency performance.

% \textbf{Shortcut on these mobile phones.}
\subsection{Shortcut verification on these mobile phones}
\label{subsec:shortcut}

Removing shortcut is concerned in CNNs architecture design recently~\cite{ding2021repvgg,vasu2023mobileone}.
To verify its necessity in light-weight CNNs, we remove the shortcut in RepGhostNet and evaluate its latency and accuracy on ImageNet. To be specific, we only remove the shortcuts of identity mapping and keep shortcut blocks for downsampling, so as to keep the model parameters and FLOPs for fair comparison. Statistically, there are 11 shortcuts of identity mapping removed in RepGhostNet. We train models with and without shortcut using the same training setting and re-parameterization structure.

% \ref{tab:shortcut} shows the accuracy and latency of RepGhostNet with and without shortcut.
We evaluate the latency performance of the networks with and without shortcuts in these mobile devices. As shown in Table~\ref{tab:shortcut}, it is clear that shortcut does not affect the actual runtime severely, but help the optimization process~\cite{he2016deep}.
% This affection is negligible for a computation-constrained mobile device.
% On one hand, thought shortcut increases memory access costs (thus affects runtime performance)~\cite{ding2021repvgg}, this affection is negligible for a computation-constrained mobile device.
On the other hand, removing shortcut of larger model (RepGhostNet 2$\times$) brings less impact on accuracy compared to smaller models, which may means that shortcut is more important to light-weight CNNs than large models, $e.g.$, RepVGG~\cite{ding2021repvgg} and MobileOne~\cite{vasu2023mobileone}.
% Besides, RepVGG~\cite{ding2021repvgg} is designed for GPU and MobileOne~\cite{vasu2023mobileone} is only evaluated with the powerful NPU in iPhone12. 
Considering all of this, we confirm that shortcut is necessary for light-weight CNNs and keep the shortcut in our RepGhostNet.

\begin{table*}[ht]
  \setlength\tabcolsep{4pt}
  \centering
  \small
  \caption{Latency and accuracy results of RepGhostNet with and without shortcut on four mobile phones. 625, 660, 845 and 970 denote the corresponding processors.}
  \begin{tabular}{cccllllllll}
    \toprule
    \multirow{2}{*}{RepGhostNet} & \multirow{2}{*}{Shortcut} & &  \multirow{2}{*}{Top1 Acc.(\%)} & & \multicolumn{5}{c}{Latency(ms)}\\ % & \(\Delta\)\\
    \cline{5-11}
    && &&  625& &660 & & 845 & & 970\\
    \midrule
    % \midrule
    \multirow{2}{*}{0.5\(\times\)}& \ding{51} & & 66.9 & 25.1& & 13.0 & & 9.0 & & 13.7\\
    & \ding{55} & & 64.1(-2.8) &24.9(-0.2)& & 12.8(-0.2) & & 8.8(-0.2) & & 13.7(-0.0)\\ % & -0.2\\
    \midrule
    \multirow{2}{*}{1.0\(\times\)}& \ding{51} & & 74.2 &62.2& & 34.2 & & 22.3 & & 32.9\\
    & \ding{55} & & 72.7(-1.5) &60.5(-1.7)& & 33.5(-0.7) & & 22.1(-0.2) & & 31.6(-1.3)\\ % & -2.0\\
    \midrule
    \multirow{2}{*}{1.5\(\times\)}& \ding{51} & & 77.5 &116.9& & 65.7 & & 42.4 & & 61.5\\
    & \ding{55} & & 75.8(-1.7) &114.9(-0.2)& & 64.8(-0.9) & & 41.9(-0.5) & & 61.4(-0.1)\\ % & -2.0\\
    \midrule
    \multirow{2}{*}{2.0\(\times\)}& \ding{51} & & 78.8 &190.0& & 105.9 & & 68.3 & & 106.4 \\
    & \ding{55} & & 77.9(-0.9) &186.7(-3.3)& & 104.3(-1.6) & & 67.5(-0.8) & & 103.8(-2.6)\\ % & -2.3\\
    \bottomrule
  \end{tabular}
  % \vspace{.25cm}
  % \vspace{-.2cm}
  \label{tab:shortcut}
  % \vspace{-.2cm}
\end{table*}

\begin{figure*}[ht]
    \centering
    \begin{subfigure}[t]{0.45\textwidth}
    \centering
   \includegraphics[trim=0cm 0cm 0cm 0cm, clip, width=1.0\linewidth]{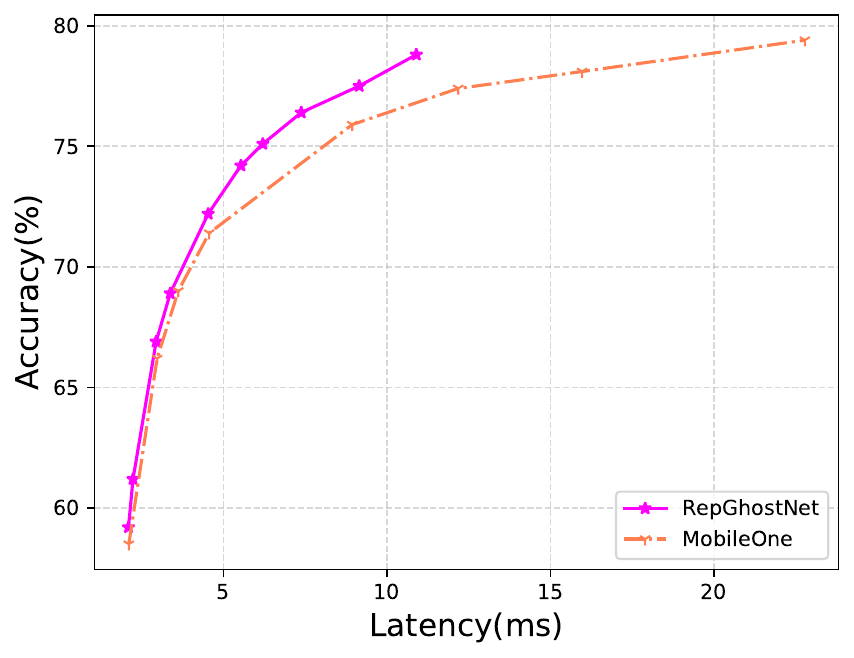}
   % \vspace{-.2cm}
   \caption{\textbf{iPhone12 CPU}}
   \label{fig:acc-to-latency-iphone12cpu}
    \end{subfigure}%
    \hfill 
    \begin{subfigure}[t]{0.45\textwidth}
        \centering
   \includegraphics[trim=0cm 0cm 0cm 0cm, clip, width=1.0\linewidth]{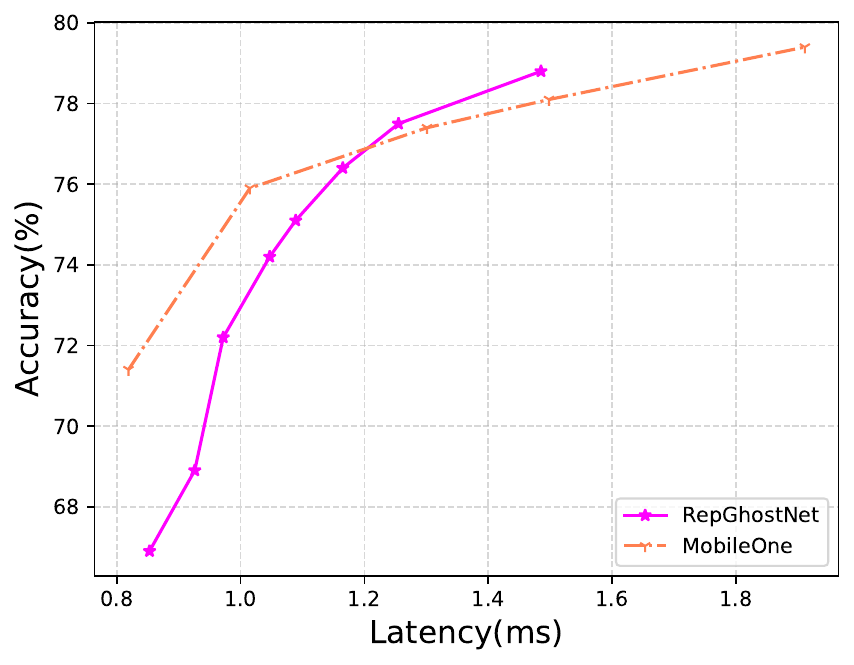}
   % \vspace{-.2cm}
   \caption{\textbf{iPhone12 NPU}}
   \label{fig:acc-to-latency-iphone12npu}
    \end{subfigure}
    \caption{Top-1 accuracy on ImageNet \textit{vs.} latency on iPhone12.}

\end{figure*}

\subsection{Comparison to MobileOne on iPhone}
\label{subsec:appendix-mobileone}
MobileOne~\cite{vasu2023mobileone} is also a recent mobile CNN, but designed for iPhone.
We provide the detailed accuracy-latency comparison of our RepGhostNet to MobileOne\footnote{\url{https://github.com/apple/ml-mobileone/tree/main/ModelBench}}~\cite{vasu2023mobileone} on the CPU and NPU\footnote{It is known as Apple Neural Engine (ANE) in iPhone, which is also a kind of Neural Processing Unit (NPU).} of an iPhone12 in Figure~\ref{fig:acc-to-latency-iphone12cpu} and Figure~\ref{fig:acc-to-latency-iphone12npu}, respectively.
While our RepGhostNet is designed for mobile CPU, it also outperforms MobileOne on iPhone12 CPU clearly as shown in Figure~\ref{fig:acc-to-latency-iphone12cpu}, same as other Android mobile phones we evaluated above.
As for iPhone12 NPU in Figure~\ref{fig:acc-to-latency-iphone12npu}, for extremely low FLOPs CNNs, RepGhostNet does not perform as well as MobileOne and RepGhostNet 1.5\(\times\) and 2.0\(\times\) outperform it. We conjecture that it is caused by the difference between CPU and NPU. With a strong parallel computing capability, NPU prefers to models with larger FLOPs than those with lower FLOPs, motivating us to design larger and more efficient CNNs for powerful NPUs in the future.

%------------------------------------------------------------------------
\section{Generalization of RepGhost}

\subsection{Generalization to MobileNetV4}
\label{subsec:repghostv2}
\begin{figure}[ht]
  \centering
  \small
  \vspace{-0.5cm}
%   \fbox{\rule{0pt}{2in} \rule{0.9\linewidth}{0pt}}
   \includegraphics[trim=0.5cm 16cm 8cm 1.5cm, clip, width=0.9\linewidth]{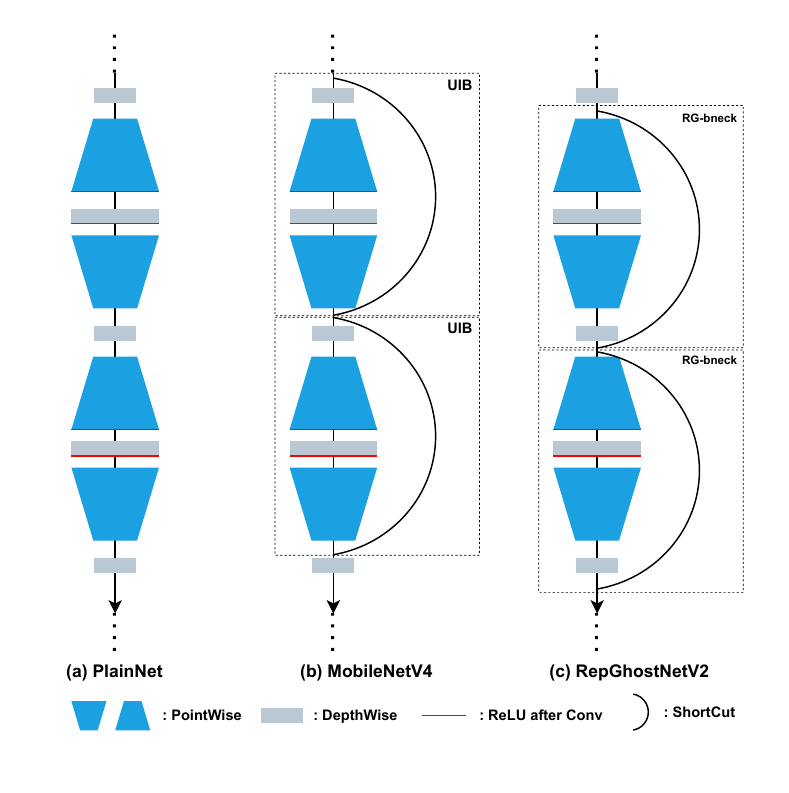}
   \vspace{-.5cm}
   \caption{Architecture comparison of MobileNetV4 and our RepGhostNetV2. (a) The plain net without shortcuts. (b) MobileNetV4. (c) Our RepGhostNetV2 has the same architecture as MobileNetV4, but only differs in the shortcut connections.}
   \label{fig:repghostv2}
   % \vspace{-.5cm}
\end{figure}

With the re-parameterization technique, we derive our novel RepGhost bottleneck in Figure~\ref{fig:repghost-bottleneck}c, whose efficiency and effectiveness are proven in the main paper. In the subsection, we use this bottleneck to evolve MobileNetV4 to our another novel network RepGhostNetV2, and also verify its architecture design again.

MobileNetV4~\cite{qin2024mobilenetv4} is a newly efficient light-weight CNNs for diverse mobile devices; it introduces a novel Universal Inverted Bottleneck (UIB) to build the efficient networks. Same as RepGhost bottleneck in Figure~\ref{fig:repghost-bottleneck}c, UIB consists of two depthwise and two pointwise convolutional layers, but differs in the layer orders, $i.e.$, '3-1-3-1' in UIB and '1-3-1-3' in ours, where 3 and 1 denote the kernel sizes and further depthwise and pointwise convolutional layers, respectively.

In this glance, when all shortcuts are removed from an UIB or RepGhost bottleneck based network, they both produce a MobileNetV1-like networks~\cite{howard2017mobilenets}, $i.e.$, alternating order of depthwise and pointwise convolutional layers like '...3-1-3-1-3-1...' layer chain without shortcuts, as PlainNet shown in Figure~\ref{fig:repghostv2}a. With this layer chain, we simply change the way of shortcut connection from UIB to our RepGhost bottleneck, thus producing our novel network in Figure~\ref{fig:repghostv2}c, which is denoted as \textit{RepGhostNetV2}.

With the same architectures except the way of shortcut connections, our RepGhostNetV2 has \textbf{exactly the same parameters and FLOPs} compared to MobileNetV4. To evaluate their performance,  we simply utilize the training setting introduced in Section~\ref{sec:exp} to both networks. Note that re-parameterization technique is not applied to RepGhostNetV2 during training. We also evaluate their latency on iPhone12 NPU.
As shown in Table~\ref{tab:acc-repghostnetv2}, RepGhostNetV2 gets similar latency on iPhone12 and comparable performance to MobileNetV4, indicating the efficiency and effectiveness of our RepGhost bottleneck. 
That is, as alternatives to UIB and MobileNetV4, our RepGhost bottleneck RepGhostNetV2 can also served as an efficient building block and light-weight CNNs for diverse mobile devices.

It is also interesting that the plain net underperforms to these two networks, but with similar latency. This proves again that the shortcuts do not affect latency but help the optimizations a lot, same as our verification in Appendix~\ref{subsec:shortcut}.

\begin{table*}[ht]
  \centering
  \small  
  \captionof{table}{Results of MobileNetV4 and RepGhostNetV2.}
    
  % \vspace{-2mm}
  \begin{tabular}{ccccccc}

    \toprule
    Model Size & Image Size & Epoch & Model & Top1 Acc(\%) & iPhone12 Latency(ms)\\
    \midrule
    \multirow{3}{*}{Small} & \multirow{3}{*}{224} & \multirow{3}{*}{300} & PlainNet & 70.7 & 0.99\\
    &&& MobileNetV4 & 72.3 & 1.01\\
    &&& RepGhostNetV2 & 72.3 & 1.01\\
    \midrule
    \multirow{3}{*}{Medium} & \multirow{3}{*}{256} & \multirow{3}{*}{300} & PlainNet & 71.6 & 1.47\\
    &&& MobileNetV4 & 79.5 & 1.48\\
    &&& RepGhostNetV2 & 79.5 & 1.48\\
    \midrule
    \multirow{3}{*}{Large} & \multirow{3}{*}{384} & \multirow{3}{*}{300}& PlainNet & 76.8 & 3.58\\
    &&& MobileNetV4 & 81.5 & 3.63\\
    &&& RepGhostNetV2 & 81.5 & 3.63\\
    \bottomrule
  \end{tabular}
    \label{tab:acc-repghostnetv2}
\end{table*}

\subsection{Generalization to YOLOv5}
To verify the replacing of concatenation to add operators, we replace the concatenation operators in $C3$ modules of YOLOv5~\cite{Jocher_YOLOv5_by_Ultralytics_2020} with add ones and keep the output channels of $C3$ modules the same.
% modify the network properly to align the parameters and FLOPs. 
% More detail will be provided in the supplementary material. 
The result in Table~\ref{tab:yolov5} shows the superior of our method.

\begin{table*}[ht]
  \centering
  \small  
  \captionof{table}{Results of YOLOv5. The models are in float16 and evaluated on a V100 GPU. Batch size are set to 32 and 1 for throughput and latency evaluations, respectively.} 
    
  % \vspace{-2mm}
  \begin{tabular}{ccccc}

    \toprule
    Model & Ops & mAP & Throughput & Latency(ms)\\
    \midrule
    \multirow{2}{*}{YOLOv5n} & Concatenation & 27.5 & 2078 & 1.31\\
    & Add & \textbf{27.6} & \textbf{2169(+4.4\%)} & \textbf{1.21(+7.3\%)}\\
    \midrule
    \multirow{2}{*}{YOLOv5s} & Concatenation & 36.1 & 1635 & 1.65\\
    & Add & \textbf{36.1} & \textbf{1757(+7.5\%)} & \textbf{1.51(+8.5\%)}\\
    \bottomrule
  \end{tabular}
    \label{tab:yolov5}
\end{table*}

\subsection{Comparison to GhostNetV2 and GhostNetV3}
GhostNetV2~\cite{tang2022ghostnetv2} augments GhostNet with DFC attention, which can be applied to improve RepGhostNet directly. Specifically, as a long-range attention, DFC is inserted after the first Ghost module of each Ghost bottleneck. Following this, we can build our RepGhostNet with DFC attention simply by inserting it after the first RepGhost module. Note that our first RepGhost module has only half the number of channels compared to that of the first Ghost module, making the attention lighter. We evaluate the latency using TFLite same as~ Section\ref{subsec:tflie}.
We also retrain GhostNetV2 using our setting and achieve accuracy similar to their reported results~\cite{tang2022ghostnetv2}.
As Table~\ref{tab:ghostnetv2_latency} shows, when equipped with DFC attention, RepGhostNet gets comparable performance to GhostNetV2 with less parameters and lower FLOPs, and even significant speedup, \(e.g.\), more than \textbf{34.0\% faster}.
% Note that we do not compare latency here because the DFC attention applied to RepGhostNet has only half the number of channels compared to that of GhostNetV2, thus introducing \textbf{lower extra latency, FLOPs and parameters}.

\begin{figure}[ht]
  \centering
  \small
  % \vspace{-.25cm}
%   \fbox{\rule{0pt}{2in} \rule{0.9\linewidth}{0pt}}
   \includegraphics[width=0.6\linewidth]{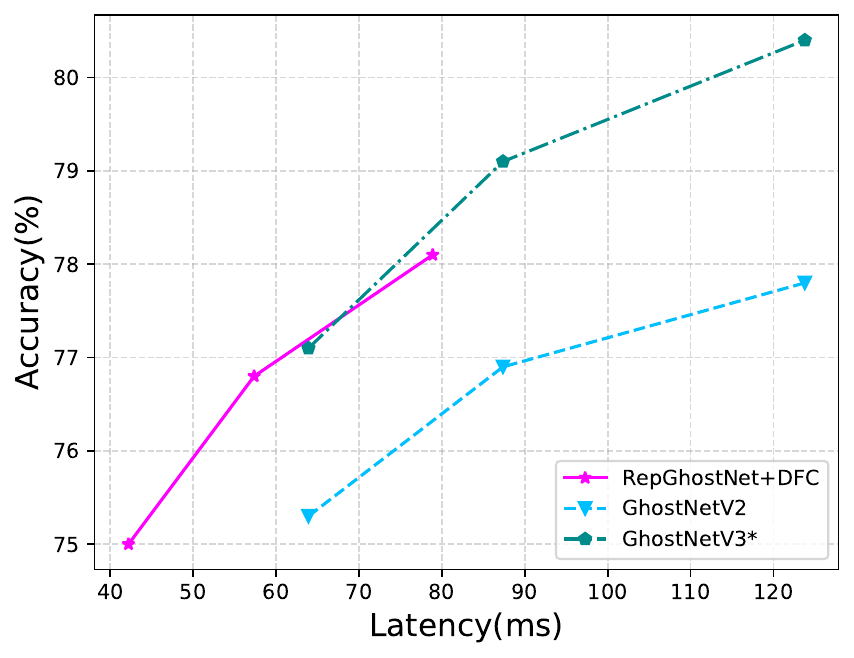}
   % \vspace{-.15cm}
   \caption{Accuracy \textbf{vs.} latency of RepGhostNet + DFC, GhostNetV2 and GhostNetV3. We retrain GhostNetV2 using our training setting. *We directly use the reported results of GhostNetV3 here, which utilizes advanced training techniques.}
   \label{fig:repghost-ghostv2-v3}
   % \vspace{-.5cm}
\end{figure}

With the same architecture to GhostNetV2, GhostNetV3~\cite{liu2024ghostnetv3} utilizes advanced training techniques, including dedicated training hyper parameters and distillation strategy, to greatly improve the performance. As shown in Figure~\ref{fig:repghost-ghostv2-v3}, our network obtains similar Pareto Frontiers to GhostNetV3 in terms of accuracy-latency trade-off, not to mention we only apply simple but effective training settings.

\begin{table*}[ht]
  % \vspace{-.3cm}
  \centering
  \small
  \caption{Comparison of RepGhostNet + DFC and GhostNetV2~\cite{tang2022ghostnetv2}.}
  \begin{tabular}{cccccc}
    \toprule
    Model & \(\alpha\) & Top1 Acc(\%) & Params(M) &  FLOPs(M) & Latency(ms)\\ 
    % \midrule                  42.19, 57.35, 78.89
    % \cellcolor{lightgray!50}  63.90, 87.37, 123.78
    \midrule
    \multirow{3}{*}{GhostNetV2} & 1.0 & \textbf{75.3} & 6.1 & 167 & 63.9\\
    & 1.3  & \textbf{76.9} & 8.9 & 269 & 87.4\\
    & 1.6  & 77.8 & 12.3 & 399 & 123.8\\
    \midrule
     RepGhostNet & 1.0 & 75.0 & \textbf{4.6} & \textbf{156} & \textbf{42.2(+34.0\%)}\\  %\multirow{3}{*}{RepGhostNet}
    + & 1.3 & 76.8 & \textbf{6.3} & \textbf{254} & \textbf{57.4(+34.3\%)}\\
    DFC & 1.6 & \textbf{78.1} & \textbf{8.4} & \textbf{370} & \textbf{78.9(+36.3\%)}\\
    \bottomrule
  \end{tabular}
  % \vspace{.001cm}
  % \vspace{-.2cm}
  \label{tab:ghostnetv2_latency}
  % \vspace{-.2cm}
\end{table*}

\begin{table*}[t]
  \centering
  \small  
  \caption{Results on comparison with Ghost-R50~\cite{han2020ghostnet}.}

  \begin{tabular}{cccccc}
    \toprule
    Model &  Params(M) & FLOPs(B) & MNN Latency(ms) & TRT Latency(ms) & Top-1(\%) \\ % & Top-5(\%)\\
    % \midrule
    \midrule
    % ResNet50\cite{he2016deep} & 25.6 & 4.1 & 778.6 & 61.5 & 77.6 & 93.7\\
    % \midrule
    Ghost-R50 (s=2) & 14.0 & 2.2 & 610.6 & 64.8 & 76.2 \\ % & 93.0\\
    RepGhost-R50 (s=1.4) & 13.2 & 2.2 & \textbf{547.1(+10.4\%)} & \textbf{56.5(+12.8\%)} & \textbf{76.2} \\ % & \textbf{93.0}\\
    \midrule
    Ghost-R50 (s=4) & 8.2 & 1.2 & 423.9 & 70.2 & 73.1 \\ % & 91.3\\
    RepGhost-R50 (s=2) & 7.1 & 1.2 & \textbf{331.6(+21.8\%)} & \textbf{38.9(+44.6\%)} & \textbf{73.6}\\ %  & \textbf{91.5}\\
    \bottomrule
  \end{tabular}
  % \vspace{-.2cm}
  \label{tab:res50_results}
  % \vspace{-.2cm}
\end{table*}

\subsection{Comparison to Ghost-R50}
To verify the generation of RepGhost module to large models, \(i.e.\), ResNet50~\cite{he2016deep}, we compare it to Ghost-R50 as reported in~\cite{han2020ghostnet}. We replace the Ghost module in Ghost-R50 with our RepGhost module to get RepGhost-R50.
All models are trained with the same training setting. MNN latency is evaluated the same as other models on the mobile device. For TRT latency, we first convert the models to TensorRT~\cite{vanholder2016efficient}, then run each model on the framework for 100 times on a T4 GPU with batch size 32, and report the average latency.
The results is shown in Table~\ref{tab:res50_results}. We can see that
RepGhost-R50 is faster than Ghost-R50 significantly in both CPU and GPU with comparable accuracy. Specially, RepGhost-R50 (s=2) gets \textbf{21.8\% and 44.6\% speedup} over Ghost-R50 (s=4) in MNN and TensorRT inferences, respectively.

%------------------------------------------------------------------------
\section{Impact Statements}
\label{sec:impact_satements}
This paper proposes a light-weight CNN architecture whose goal is to improve the efficiency of CNNs for mobile devices, \(i.e.\), less computational resources and higher accuracy. With this positive impact, we believe our model will make deep learning on computer vision much more widely and easily accessible, especially for people from less developed areas.

As a data-driven deep learning architecture, our RepGhostNet compares to other state-of-the-art light-weight CNNs using public datasets in our work, like ImageNet and COCO, which verify its efficiency and effectiveness. However, training our models using other datasets may lead to some dataset-related impacts which is beyond our scope and also faced by all other deep learning architectures. We leave it for future works.

\end{document}